\documentclass{article}
\usepackage{arxiv}
\usepackage{amsfonts}
\usepackage{amsmath}
\usepackage{amssymb}
\usepackage{bm}
\usepackage{todonotes}
\usepackage{microtype}
\usepackage{graphicx}
\usepackage{booktabs}
\usepackage{tikz}
\usepackage{epstopdf}
\usepackage{algpseudocode}
\usepackage{algorithm}
\usepackage{orcidlink}
\usepackage{enumitem}
\usepackage{subcaption}
\usepackage{multirow}
\usepackage{bbm}
\usepackage{biblatex}
\usepackage{cleveref}

\makeatletter
\newrobustcmd*{\parentexttrack}[1]{%
  \begingroup
  \blx@blxinit \blx@setsfcodes \blx@bibopenparen#1\blx@bibcloseparen \endgroup}
  \DeclareUnicodeCharacter{03BC}{$\mu$} \AtEveryCite{%
  }
\makeatother

\setlist[enumerate]{leftmargin=.5in}
\setlist[itemize]{leftmargin=.5in}
\ifpdf
  \DeclareGraphicsExtensions{.eps,.pdf,.png,.jpg}
\else
  \DeclareGraphicsExtensions{.eps}
\fi

\crefformat{equation}{Eq.~(#2#1#3)}

\DeclareMathOperator*{\argmin}{arg\,min}
\DeclareMathOperator*{\argmax}{arg\,max}
\newcommand{\fb}{\mathbf{f}}\newcommand{\Fb}{\mathbf{F}}\newcommand{\yb}{\mathbf{y}}
\newcommand{\gb}{\mathbf{g}}\newcommand{\Ex}{\mathbb{E}}\newcommand{\Prob}{\mathbb{P}}
\newcommand{\Cov}{\mathrm{Cov}} \newcommand{\Xspace}{\mathcal{X}}
\newcommand{\Uspace}{\mathcal{U}} \newcommand{\ParFront}{\mathcal{P}^*}
\newcommand{\Xoptim}{\mathfrak{X}} \newcommand{\ParSet}{\mathcal{P}^*_\Xspace}
\newcommand{\mb}{\mathbf{m}} 
\renewcommand{\epsilon}{\varepsilon}

\newcommand{\HV}{\mathop{\mathrm{HV}}} \newcommand{\UHV}{\mathop{\mathrm{UHV}}}
\newcommand{\Action}{\mathcal{A}}

\title{Expected Hypervolume Maximization for Multiobjective Optimization under Uncertainties}
\date{}
\author{Victor Trappler\,\orcidlink{0000-0003-4620-4861} \\Departement GMI \\Institut Henri Fayol,
 \\Mines Saint-Etienne, Univ Clermont Auvergne,\\CNRS, UMR 6158 LIMOS\\ F - 42023 Saint-Etienne France
}

\begin{document}

\maketitle


\begin{abstract}
The problem of multiobjective optimization under uncertainties is often
approached by taking the expectation of each objective. In this work, we propose
instead to formulate this as a Bayesian decision problem and to rely on the
expected value of the hypervolume, which is to be maximized with respect to a
finite set of input points. We show that this can be performed using methods
based on gradients in a stochastic optimization framework, provided that care is
taken with respect to dominated points.
Moreover, in the absence of readily available
differentiable code, we propose to use Gaussian Processes as differentiable
surrogate models, in order to perform the optimization. An additional
contribution in this work are some active learning strategies, through
acquisition functions which helps construct a surrogate model well-designed for
the multiobjective optimization problem at stake. These strategies are compared
on simple analytical problems to assess their performances.

\end{abstract}



\section{Introduction}
\subsection{Context}
In many scientific and industrial fields, the decision-making process relies on
the optimization of several objectives. However, in most practical cases, those
objectives are concurrent, as they do not share the same optimizers.
Various approaches have been suggested in order to aggregate all the objective,
by means of scalarization methods such as linear combinations, or variations
around Tchebycheff scalarization \cite{lin_smooth_2024a}. A more general
approach relies on the definition of a preorder in the objective space, which
allows defining the set of all possible optimal tradeoffs between the
objectives, which is the Pareto front, and its pre-image the Pareto set.

Various methods can be found in the literature in order to evaluate the Pareto
front and Pareto set, using for instance evolutionary algorithm such as NSGA-II
\cite{deb_fast_2002}. With the advent of differentiable programming and
autodifferentiation tools, gradient-based methods have also been proposed
\cite{wang_hypervolume_2017} while for expensive-to-evaluate objective function,
Bayesian Optimization has also been widely used \cite{emmerich_single_2006}.

However in real-world optimization problems, the objectives are affected in some
way by uncertainties, leading to the definition of Stochastic Multiobjective
Optimization Problems (SMOOUU), sometimes called multi-objective stochastic
optimization problems (MOSOPs) \cite{pirouz_multiobjective_2026}. Different ways
have been proposed to tackle the multiobjective optimization problem under
uncertainties. Some authors extend the domination preorder to the probabilistic
setting by introducing a threshold to consider that a solution dominates
another with high enough probability
\cite{congedo_multiobjective_2015,mores_multiobjective_2023,laag_centeroutward_2025},
while others propose to optimize statistical quantities of the objectives
\cite{basseur_handling_2006,daulton_robust_2022,pirouz_multiobjective_2026}.
\subsection{Contributions}
We summarize our main contributions as follows:
\begin{itemize}
    \item We propose to frame the SMOOUU problem as a Bayesian decision problem
using the hypervolume as an utility function, and where the optimization is
performed with respect to a finite set of candidate, as shown in
\cref{sec:bayesian_optimal_decision}.
\item We show how the optimization can be carried using stochastic gradient
methods in \cref{sec:moo_uhv} using the notion
of Uncrowded hypervolume to account for null-gradients
\item Finally, because of the
computational cost associated with this process, we propose to use surrogate
models based on Gaussian Processes as proxies of the function in order to ease
the computational effort needed to perform this optimization, along with an
active learning procedure to improve the surrogate model with respect to the
task at hand in \cref{sec:gp_bo} and \cref{sec:lookahead}, in the case of a
simulator setting.
\end{itemize}

\subsection{Related work}
There have been some work on the way to quantify uncertainties around Pareto
Fronts, either by considering them as random sets \cite{binois_quantifying_2015}
or by using a polar parameterization in \cite{tu_random_2024}. As mentioned
earlier, the Pareto preorder can be extended to take into account uncertainties
and allow for comparing stochastic vectors for an optimization context
\cite{khosravi_probabilistic_2018,mores_multiobjective_2023,congedo_multiobjective_2015}
which allows for constructing an optimization procedure. This usually requires
setting a threshold to assess whether the probability of domination is large
enough.
Another approach is to focus on the notion "optimality" under uncertainties:
\cite{daulton_robust_2022} uses the notion of Multivariate Value-at-Risk, while
\cite{trappler_multiobjective_2025} look to maximize the probability of being
Pareto Optimal, allowing to rank solutions in the face of uncertainties.

For practical applications, some aspects of Bayesian Optimization have been
applied to this problem, but either focus on input noise perturbations
\cite{daulton_robust_2022}, output perturbations as in \cite{han_novel_2022}, or
propose extensions of commonly used acquisition functions in the deterministic
case \cite{trappler_multiobjective_2025} for specific objectives. The case of
stochastic simulators, that is when the uncertainty is directly embedded in the
evaluation of the objectives, has also been studied extensively in the
literature, as in
\cite{rojasgonzalez_multiobjective_2020,pal_multiobjective_2020}.

In~\cite{basseur_handling_2006}, the authors propose
a method directly related to the one we introduce here, where they use the
expected value of the $\epsilon$-indicator to define a fitness function which is
used within an evolutionary algorithm.

\section{Stochastic Multiobjective Optimization as a Bayesian decision problem}
\label{sec:bayesian_optimal_decision}
\subsection{Bayesian optimal decision under uncertainties}
Let $(\Omega, \mathcal{F}, \Prob)$ be a probability space. We can consider the
optimization problem under uncertainties as a Bayesian decision problem by
defining a generic loss function
\begin{equation}
    L(\omega, a): \Omega \times \Action \rightarrow \mathbb{R}\,,
\end{equation}
where $a \in \Action$ is the action or decision to be taken, and
$\omega\in\Omega$ represents the state of nature. This loss quantifies the cost
of taking the decision $a$, when nature takes the value $\omega$.

The Bayes Optimal decision $a_{\text{Bayes}}$ minimizes Bayes' risk, which is
defined as the expectation with respect to the measure defined on $\Omega$:
\begin{align}
\label{eq:bayes_opt_decision}
    a_{\text{Bayes}} \in \argmin_{a\in\Action}\Ex\left[L(\omega, a)\right].
    \end{align}
If the uncertainty in the problem is modelled using a random variable $U$ with
sample space $\Uspace$ and known probability distribution function $p_U$, we can
rewrite the integral above as
\begin{align} 
    a_{\text{Bayes}}&\in \argmin_{a \in \Action} \int L(\omega, a)\,\mathrm{d}\omega= \argmin_{a \in \Action} \int_{\Uspace}\ell(u, a)p_U(u)\,\mathrm{d}u\,,
\end{align}
with $ L(\omega, \cdot)=\ell(U(\omega), \cdot)$.

In the usual problem of minimization under uncertainties, we are looking to
optimize an objective function, which is affected by a environmental parameter:
$f: \Xspace \times \Uspace \rightarrow \mathbb{R}$. In this case, the action
space $\Action$ is the search space $\Xspace$, the loss can be expressed simply
as the value taken by the function,
\begin{equation}
    L(\omega, x) = f(x, U(\omega)).
\end{equation}
According to \cref{eq:bayes_opt_decision} the Bayesian optimal
decision for this problem is
\begin{equation}
     x_{\text{Bayes}} \in \argmin_{x\in\Xspace}\Ex_U\left[f(x, U)\right],
\end{equation}
so the Bayesian optimal solution is to minimize the mean of the function.

In multiobjective optimization, the objective function is actually a
concatenation of multiple individual objectives. Let $\Xspace \subset
\mathbb{R}^{n_x}$, and let the objective function
\begin{equation}
\begin{array}{rcl}
\fb: \Xspace \times \Uspace& \longrightarrow& \mathbb{R}^d \\
x & \longmapsto &\fb(x,u) = (f_1(x,u),\dots,f_d(x,u))
\end{array}.
\end{equation}
In this case, without uncertainties, we are looking for a subset of the input
space $\Xspace$ which gives the non-dominated points in the objective space.
We will focus now on the \emph{simulator setting}, where the function $\fb$ is
deterministic, meaning that for any $x,u$, $\fb(x, u)$ is deterministic.
\subsection{Loss function in multiobjective optimization under uncertainties}
\subsubsection{Pareto Dominance and Indicator functions}

Comparing vectors of function evaluation can be done by introducing a partial
order on $\mathbb{R}^d$. For $\yb = (y_1,\dots, y_d)$ and $\yb' =
(y'_1,\dots,y'_d)$, we say that $\yb$ dominates $\yb'$, and we note $\yb \prec
\yb'$ if $\forall i,\, y_i \leq y_i' \text{ and } \exists j $ such that $y_j
<y_j'$. In other words, $\yb \prec \yb'$ means that $\yb$ is at least as good as
$\yb'$ in all objectives (for a multiobjective minimization problem), and
strictly better in at least one. Those relations can be used to compare sets of
vectors which can act as Pareto front approximations. A short summary is
displayed \cref{tab:pareto_domination}, using notations from
\cite{audet_performance_2021}.
\begin{table}
    \begin{center}
        \begin{tabular}{rll}\toprule
           & Notation& Definition \\\midrule
           \multirow{2}{*}{Point comparison}& $\yb \preceq \yb'$ & $y_i \leq y_i'$ for all $i$\\
            &$\yb \prec \yb'$ &  $y_i \leq y_i'$ for all $i$ and $\exists j$ such
            that $y_j < y_j'$\\ \midrule
             \multirow{3}{*}{Set comparison}&$A \preceq B$ & For any $\yb_B \in B$, $\exists \yb_A \in A$,  $\yb_A
            \preceq \yb_B$ \\
            &$A \prec B$ & For any $\yb_B \in B$, $\exists \yb_A \in A$,  $\yb_A
            \prec \yb_B$ \\
            &$A \vartriangleleft B$ & For any $\yb_B \in B$, $\exists \yb_A \in A$,  $\yb_A
            \preceq \yb_B$ and $A\neq B$  \\\bottomrule
        \end{tabular}   
    \end{center}
    \caption{Short summary of notation used for the dominance relation}
    \label{tab:pareto_domination}
\end{table}

Those Pareto front approximations can also be measured quantitatively, by
introducing a measure of the quality of the approximation as an unary indicator
$I: P(\mathcal{Y})\to \mathbb{R}$, which maps a subset of the objective space to
a real value. Different unary indicators can be defined
\cite{zitzler_performance_2003,zitzler_quality_2008}, such as the cardinality of
the empirical Pareto front, or the hypervolume.
 One desirable property for a unary indicator is the monotonicity
\cite{zitzler_quality_2008}: let us consider two approximations of the Pareto Front, say
$\mathcal{Y}_1$ and $\mathcal{Y}_2$. If $\mathcal{Y}_2$ is weakly dominated by
$\mathcal{Y}_1$ (every  point of $\mathcal{Y}_2$ is dominated by at least one of
$\mathcal{Y}_1$), then an indicator $I$ is monotonic (assuming a greater value
is preferable) if it verifies
\begin{equation}
    \mathcal{Y}_1 \preceq \mathcal{Y}_2 \Rightarrow I(\mathcal{Y}_1) \geq I(\mathcal{Y}_2)\,.
\end{equation}

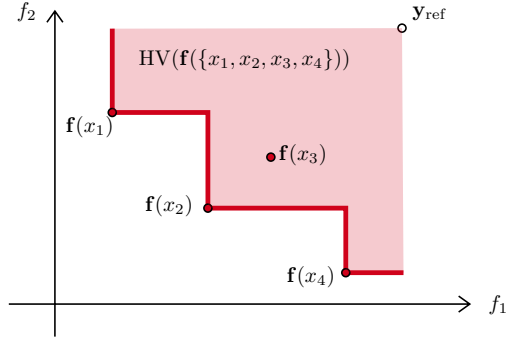
\begin{figure}
    \centering
    \scalebox{0.8}{\tikzset{every picture/.style={line width=0.75pt}} 

\begin{tikzpicture}[x=0.75pt,y=0.75pt,yscale=-1,xscale=1]

\draw [color={rgb, 255:red, 208; green, 2; blue, 27 }  ,draw opacity=1 ][line width=2.25]    (210.9,36) -- (210.9,88.6) -- (270.9,88.6) -- (270.9,148.6) -- (357.4,148.6) -- (357.4,189.1) -- (393.6,189.1) ;
\draw  (146,208.78) -- (435.2,208.78)(174.92,25) -- (174.92,229.2) (428.2,203.78) -- (435.2,208.78) -- (428.2,213.78) (169.92,32) -- (174.92,25) -- (179.92,32)  ;
\draw  [fill={rgb, 255:red, 208; green, 2; blue, 27 }  ,fill opacity=1 ] (208.4,88.6) .. controls (208.4,87.22) and (209.52,86.1) .. (210.9,86.1) .. controls (212.28,86.1) and (213.4,87.22) .. (213.4,88.6) .. controls (213.4,89.98) and (212.28,91.1) .. (210.9,91.1) .. controls (209.52,91.1) and (208.4,89.98) .. (208.4,88.6) -- cycle ;
\draw  [fill={rgb, 255:red, 208; green, 2; blue, 27 }  ,fill opacity=1 ] (268.4,148.6) .. controls (268.4,147.22) and (269.52,146.1) .. (270.9,146.1) .. controls (272.28,146.1) and (273.4,147.22) .. (273.4,148.6) .. controls (273.4,149.98) and (272.28,151.1) .. (270.9,151.1) .. controls (269.52,151.1) and (268.4,149.98) .. (268.4,148.6) -- cycle ;
\draw  [fill={rgb, 255:red, 208; green, 2; blue, 27 }  ,fill opacity=1 ] (354.9,189.1) .. controls (354.9,187.72) and (356.02,186.6) .. (357.4,186.6) .. controls (358.78,186.6) and (359.9,187.72) .. (359.9,189.1) .. controls (359.9,190.48) and (358.78,191.6) .. (357.4,191.6) .. controls (356.02,191.6) and (354.9,190.48) .. (354.9,189.1) -- cycle ;
\draw   (390.17,35.67) .. controls (390.17,34.29) and (391.29,33.17) .. (392.67,33.17) .. controls (394.05,33.17) and (395.17,34.29) .. (395.17,35.67) .. controls (395.17,37.05) and (394.05,38.17) .. (392.67,38.17) .. controls (391.29,38.17) and (390.17,37.05) .. (390.17,35.67) -- cycle ;
\draw  [draw opacity=0][fill={rgb, 255:red, 208; green, 2; blue, 27 }  ,fill opacity=0.21 ] (392.67,35.67) -- (393.6,189.1) -- (357.4,189.1) -- (357.4,148.6) -- (270.9,148.6) -- (270.9,88.6) -- (210.9,88.6) -- (210.9,36) -- cycle ;
\draw  [fill={rgb, 255:red, 208; green, 2; blue, 27 }  ,fill opacity=1 ] (307.9,116.6) .. controls (307.9,115.22) and (309.02,114.1) .. (310.4,114.1) .. controls (311.78,114.1) and (312.9,115.22) .. (312.9,116.6) .. controls (312.9,117.98) and (311.78,119.1) .. (310.4,119.1) .. controls (309.02,119.1) and (307.9,117.98) .. (307.9,116.6) -- cycle ;

\draw (445.4,201.7) node [anchor=north west][inner sep=0.75pt]    {$f_{1}$};
\draw (149.5,17.9) node [anchor=north west][inner sep=0.75pt]    {$f_{2}$};
\draw (397.83,20.83) node [anchor=north west][inner sep=0.75pt]   [align=left] {$\mathbf{y}_{\mathrm{ref}}$};
\draw (180.17,89.33) node [anchor=north west][inner sep=0.75pt]   [align=left] {$\displaystyle \fb( x_{1})$};
\draw (230,139) node [anchor=north west][inner sep=0.75pt]   [align=left] {$\displaystyle \fb( x_{2})$};
\draw (319,183.33) node [anchor=north west][inner sep=0.75pt]   [align=left] {$\displaystyle \fb( x_{4})$};
\draw (226.17,47.07) node [anchor=north west][inner sep=0.75pt]    {$\mathrm{HV}(\mathbf{f}(\{x_{1} ,x_{2} ,x_{3} ,x_{4}\}))$};
\draw (314,107) node [anchor=north west][inner sep=0.75pt]   [align=left] {$\displaystyle \fb( x_{3})$};

\end{tikzpicture}}
    \caption{Hypervolume of the region dominated by $\fb(\{x_1, x_2, x_3,
    x_4\})$, upper bounded by $\yb_{\mathrm{ref}}$ in shades of red.}
    \label{fig:hypervolume_sketch}
\end{figure}

A well-known unary indicator function is the hypervolume of the region dominated
by the points of the approximation \cite{falcon-cardona_construction_2019},
which can then be expressed for an approximation $\mathcal{Y}$ as 
\begin{equation}
    \HV(\mathcal{Y}) = \int_{\mathbb{B}_{\text{ref}}} \boldsymbol{1}_{\{\mathcal{Y} \prec \yb\}} \, \mathrm{d}\yb \,.
\end{equation}
Here $\mathbb{B}_{\text{ref}}$ is defined as $\{ \yb \mid \yb \prec
\yb_{\text{ref}}\}$, and is necessary to bound the region using a reference
point $\yb_{\text{ref}}$, and $\boldsymbol{1}_{\{\mathcal{Y} \prec \yb\}} = 1$ if
$\yb$ is dominated by $\mathcal{Y}$, $0$ elsewhere. The choice of
$\yb_{\text{ref}}$ is crucial as it should be dominated by every point of the
true Pareto set, so that the hypervolume attains its maximum for the true Pareto
front. An illustration of the hypervolume can be found
\cref{fig:hypervolume_sketch}, where $\mathcal{Y} = \fb(\{x_1, x_2, x_3,
x_4\})$. In this case, $\fb(x_3)$ does not contribute to the hypervolume. Since
it is monotonic, the maximal value of the hypervolume is attained by the true
Pareto front, provided that $\yb_{\text{ref}}$ is dominated by every point in
this Pareto front.  In what follows, we will assume that $\yb_{\text{ref}}$ is
chosen to verify this, and the dependence of the hypervolume to the reference
point will be omitted.

Using the hypervolume, we can look for the ``best'' discrete
approximation of the Pareto front using $n$ points $\ParFront_{n}$
\cite{auger_theory_2009,auger_theoretically_2010} by looking for a solution to
the optimization problem defined as
\begin{equation}
    \max_{\substack{\Xoptim \subset \Xspace \\|\Xoptim| \leq n} }\HV(\fb(\Xoptim)) \label{eq:parfront_HV_n}\,,
\end{equation}
where the maximization problem is with respect to sets of at most $n$ points of
$\Xspace$.

\subsubsection{Bayesian optimal decision for SMOOUU}
Based on the definition of the optimal approximation of the Pareto front
\cref{eq:parfront_HV_n}, we can reframe the SMOOUU problem as a Bayesian decision
problem as introduced \cref{eq:bayes_opt_decision}.

The action space is then $\Action = \Xspace^n$, and the loss can be chosen as the opposite of the hypervolume:
\begin{equation}
    L(\omega, \Xoptim) = -\HV(\fb\left(\Xoptim, U(\omega)\right)) = -\HV(\{\fb(x, U(\omega)) \text{ for } x\in \Xoptim\})\,,
\end{equation}
thus the Bayesian optimal decision is now a vector of $n$ elements of $\Xspace$
such that
\begin{align}
\mathfrak{X}_{\mathrm{Bayes}} &\in\argmax_{\Xoptim \in \Action} \int \HV(\fb(\Xoptim, U(\omega)))\,\mathrm{d}\omega \nonumber\\
   &\in \argmax_{\Xoptim \in \Xspace^n} \Ex_U\left[\HV(\fb(\Xoptim, U))\right] \label{eq:bayes_expected_hv} \,.
\end{align}

The actual optimization of this quantity is treated using gradient-based
optimization in \cref{sec:moo_uhv}. We will focus now on different properties of
this quantity.

\subsection{Properties of the expected hypervolume}
\subsubsection{Single candidate}
In the simplest case, we consider a single candidate, meaning that we set $n=1$.
The hypervolume can be rewritten simply as
\begin{equation}
    \HV(\fb(x, U)) = \prod_{i=1}^d ({r}_i - f_i(x, U))\,.
\end{equation}
For the two objective case, $d=2$:
\begin{align}
    \HV(\fb(x,u))  = (r_1 - f_1(x,u))\cdot (r_2 - f_2(x,u))\,,
\end{align}
and taking the expectation with respect to $U$ gives
\begin{align}
        \Ex_U\left[\HV(\fb(x,U))\right] &=  \mathrm{Cov}(f_1(x, U), f_2(x, U)) + \HV(\Ex_U\left[\fb(x, U)\right])\,,
\end{align}
which highlights the role of the covariance between the objectives. In this
case, a positive covariance between the objectives will have a tendency to
increase the expected hypervolume. When considering three objectives ($d=3$), a
central quantity is the third order mixed cumulant, the coskewness, defined as
\begin{align}
    \mathrm{Coskew}(X,Y,Z) = \Ex\left[\left(X-\Ex[X]\right)\left(Y-\Ex[Y]\right)\left(Z-\Ex[Z]\right)\right]\,,
\end{align}
which is positive when either the three of them are having deviations
simultaneously in the positive direction, or when two of them are having
deviations in the negative direction, while the remaining one toward the
positive direction. Moreover, we have for any $a,b,c \in\mathbb{R}$, 
$\mathrm{Coskew}(X+a,Y+b,Z+c)  = \mathrm{Coskew}(X,Y,Z)$ and
$\mathrm{Coskew}(aX,bY,cZ)  = (abc)\cdot\mathrm{Coskew}(X,Y,Z)$

Writing out the expected value of the hypervolume for the $d=3$ case reads
\begin{align}
    \Ex_U\left[\HV(\fb(x,U))\right] = 
  &\HV(\Ex_U\left[\fb(x,U)\right])- \mathrm{Coskew}(f_1(x,U),f_2(x,U),f_3(x,U))\\ \nonumber
    & +\Cov(f_1(x,U),f_2(x,U))\Ex_U\left[r_3-f_3(x,U)\right] \\ \nonumber
    & +\Cov(f_2(x,U),f_3(x,U))\Ex_U\left[r_1-f_1(x,U)\right] \\ \nonumber
        & +\Cov(f_3(x,U),f_1(x,U))\Ex_U\left[r_2 - f_2(x,U)\right] \nonumber\,.
\end{align}
Once again, we can see that positive covariance between the objectives will
have a positive impact on the expected value of the hypervolume. However, the
coskewness here has a negative impact, meaning that the expected hypervolume
will decrease when the three objectives improve simultaneously.

\subsubsection{Multiple candidates with two objectives}
We now consider the general case of $n$ designs taken simultaneously
$\{x_i\}_{1\leq i \leq n}$, with $d=2$ objectives. Without loss of generality,
we consider that the points $\{x_i\}_{1\leq i \leq n}$ are sorted according to
the first objective in increasing order, that is $f_1(x_1)< f_1(x_2) <\dots <
f_1(x_n)$.

\begin{figure}
    \centering
    \scalebox{.6}{\tikzset{every picture/.style={line width=0.75pt}} 

\begin{tikzpicture}[x=0.75pt,y=0.75pt,yscale=-1,xscale=1]

\draw [color={rgb, 255:red, 208; green, 2; blue, 27 }  ,draw opacity=1 ][line width=2.25]    (210.9,36) -- (210.9,88.6) -- (270.9,88.6) -- (271.4,127.2) -- (357.4,127.2) -- (357.4,189.1) -- (393.6,189.1) ;
\draw  (146,208.78) -- (435.2,208.78)(174.92,25) -- (174.92,229.2) (428.2,203.78) -- (435.2,208.78) -- (428.2,213.78) (169.92,32) -- (174.92,25) -- (179.92,32)  ;
\draw  [draw opacity=0][fill={rgb, 255:red, 208; green, 2; blue, 27 }  ,fill opacity=1 ] (205.71,88.6) .. controls (205.71,85.73) and (208.03,83.41) .. (210.9,83.41) .. controls (213.77,83.41) and (216.09,85.73) .. (216.09,88.6) .. controls (216.09,91.47) and (213.77,93.79) .. (210.9,93.79) .. controls (208.03,93.79) and (205.71,91.47) .. (205.71,88.6) -- cycle ;
\draw   (390.17,35.67) .. controls (390.17,34.29) and (391.29,33.17) .. (392.67,33.17) .. controls (394.05,33.17) and (395.17,34.29) .. (395.17,35.67) .. controls (395.17,37.05) and (394.05,38.17) .. (392.67,38.17) .. controls (391.29,38.17) and (390.17,37.05) .. (390.17,35.67) -- cycle ;
\draw  [draw opacity=0][fill={rgb, 255:red, 208; green, 2; blue, 27 }  ,fill opacity=0.21 ] (392.67,35.67) -- (393.6,189.1) -- (357.4,189.1) -- (357.4,127.2) -- (271.4,127.2) -- (270.9,88.6) -- (210.9,88.6) -- (210.9,36) -- cycle ;
\draw  [draw opacity=0][fill={rgb, 255:red, 208; green, 2; blue, 27 }  ,fill opacity=1 ] (266.21,127.2) .. controls (266.21,124.33) and (268.53,122.01) .. (271.4,122.01) .. controls (274.27,122.01) and (276.59,124.33) .. (276.59,127.2) .. controls (276.59,130.07) and (274.27,132.39) .. (271.4,132.39) .. controls (268.53,132.39) and (266.21,130.07) .. (266.21,127.2) -- cycle ;
\draw  [draw opacity=0][fill={rgb, 255:red, 208; green, 2; blue, 27 }  ,fill opacity=1 ] (352.21,189.1) .. controls (352.21,186.23) and (354.53,183.91) .. (357.4,183.91) .. controls (360.27,183.91) and (362.59,186.23) .. (362.59,189.1) .. controls (362.59,191.97) and (360.27,194.29) .. (357.4,194.29) .. controls (354.53,194.29) and (352.21,191.97) .. (352.21,189.1) -- cycle ;
\draw  [draw opacity=0][fill={rgb, 255:red, 126; green, 211; blue, 33 }  ,fill opacity=1 ] (270.9,82.04) -- (277.46,88.6) -- (270.9,95.16) -- (264.34,88.6) -- cycle ;
\draw  [draw opacity=0][fill={rgb, 255:red, 126; green, 211; blue, 33 }  ,fill opacity=1 ] (357.4,120.64) -- (363.96,127.2) -- (357.4,133.76) -- (350.84,127.2) -- cycle ;

\draw (445.4,201.7) node [anchor=north west][inner sep=0.75pt]    {$f_{1}$};
\draw (149.5,17.9) node [anchor=north west][inner sep=0.75pt]    {$f_{2}$};
\draw (407.83,16.83) node [anchor=north west][inner sep=0.75pt]   [align=left] {nadir};

\end{tikzpicture}}
    \caption{Hypervolume of the region dominated by three points. The red circles mark the vectors where a positive correlation increases the expected hypervolume, while the green rhombuses indicate where a positive correlation decreases it}
    \label{fig:HV_2x2}
\end{figure}
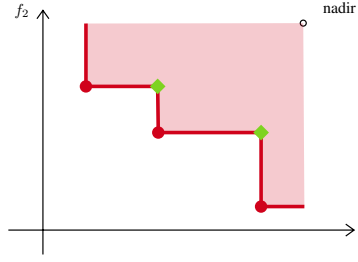

For $n=2$, one simplification we will make is to consider that all designs
stay Pareto optimal almost surely.
Given the assumptions that $\fb(x_1, U)$ and $\fb(x_2, U)$ are non-dominated
almost surely, it is straightforward to show that
\begin{align}
    \Ex_U\left[\HV\left(\fb(\{x_1,x_2\}, U)\right)\right] \label{eq:EHV_2d_2}
    = &\HV\left(\Ex_U[\fb(\{x_1,x_2\}, U)]\right) -\Cov(f_1(x_2,U),f_2(x_1,U))  \\ &+ \Cov(f_1(x_1,U),f_2(x_1, U)) + \Cov(f_1(x_2,U),f_2(x_2,U))  \nonumber\,,
\end{align}
and more generally, when all the $\fb(\{x_i\}_i, U)$ stay non-dominated with
respect to each other, we have that
\begin{multline}
        \Ex_U\left[\HV(\fb(\{x_i\}_i, U)\right]= \HV\left(\Ex_U\left[\fb(\{x_i\}, U)\right]
        \right) \\+ \sum_{i=1}^n \Cov(f_1(x_i,U),f_2(x_i,U)) - \sum_{i=2}^n\Cov(f_1(x_i,U),f_2(x_{{i-1}},U))\,.
        \label{eq:EHV_2d_n}
\end{multline}
\Cref{eq:EHV_2d_2,eq:EHV_2d_n} show that when the objectives are pairwise non
correlated, the expected value of the hypervolume is equal to the hypervolume of
the expected objective. This is the case when the uncertainty is represented as
noise added to the objectives, and where each objective is affected
independently. As before, a positive covariance between
objectives increases the expected hypervolume, as in the single candidate case,
but a positive covariance term between the two objectives and successive
candidates (which are the inward points of the region) decreases the expected
hypervolume. This is illustrated \cref{fig:HV_2x2}.

\subsubsection{Upper bound}
The maximal value of the expected hypervolume can be upper bounded using the
whole input space:
\begin{align}
    \max_{\mathfrak{X}\in\Xspace^n}\Ex_U\left[\HV\left(\fb(\mathfrak{X}, U\right)\right] \leq \Ex_U\left[\max_{\mathfrak{X}\in \Xspace^n} \HV\left(\fb(\mathfrak{X}, U)\right)\right] &\leq \Ex_U\left[\max_{\mathfrak{X}\subset \Xspace} \HV(\fb(\mathfrak{X},U))\right] \\ 
    &= \Ex_U\left[\HV(\fb(\Xspace, U))\right]\label{eq:hv_ub}\,,
\end{align}
where $\fb(\Xspace, U)$ is the image of the whole control space under
$\fb(\cdot, U)$, and the last equality stems from the fact that only the non
dominated points of $\fb(\Xspace, U)$ contributes to the hypervolume, or in
other words, that adding dominated points does not reduce the hypervolume.

Increasing the number of points considered in the optimization $n$ reduces the
gap between the left-hand side which is the optimum and the upper bound,
obtained using the whole control space. In what follows, we will refer to the
difference between the upper bound $\Ex_U\left[\HV(\fb(\Xspace, U))\right]$ and the
optimal value (the left-hand side of  the inequality \cref{eq:hv_ub}) as the
optimization gap. The gap is illustrated later on
\cref{fig:inequality_gap_truth} for a simple problem.

Using the notion of conditional Pareto sets introduced in
\cite{trappler_multiobjective_2025}, which defines for $u\in\Uspace$ the Pareto
set of $\fb(\cdot, u)$ written $\ParSet(u)$, we have by definition that this
conditional Pareto set maximizes the HV 
\begin{equation}
    \HV(\fb(\Xspace), u) = \HV(\fb(\ParSet(u), u))\,,
\end{equation}
thus taking the union of all the conditional Pareto sets allow us to reach the
upper bound as well:
\begin{equation}
     \Ex_U\left[\HV(\fb(\Xspace), U) \right]= \Ex_U\left[\HV\left(\fb\left(\bigcup_{\tilde{u}\in\Uspace}\ParSet(\tilde{u}), U\right)\right)\right]\,.
\end{equation}
This suggests that the optimal set of $n$ control points $\mathfrak{X}\in
\Xspace^n$ which maximizes the expected hypervolume should cover at least
partially the union of the conditional Pareto sets.

\section{Expected Uncrowded hypervolume}
\label{sec:moo_uhv}
\subsection{Uncrowded hypervolume for deterministic MOO}

\begin{figure}[!ht]
    \centering
    \includegraphics[width=0.6\linewidth]{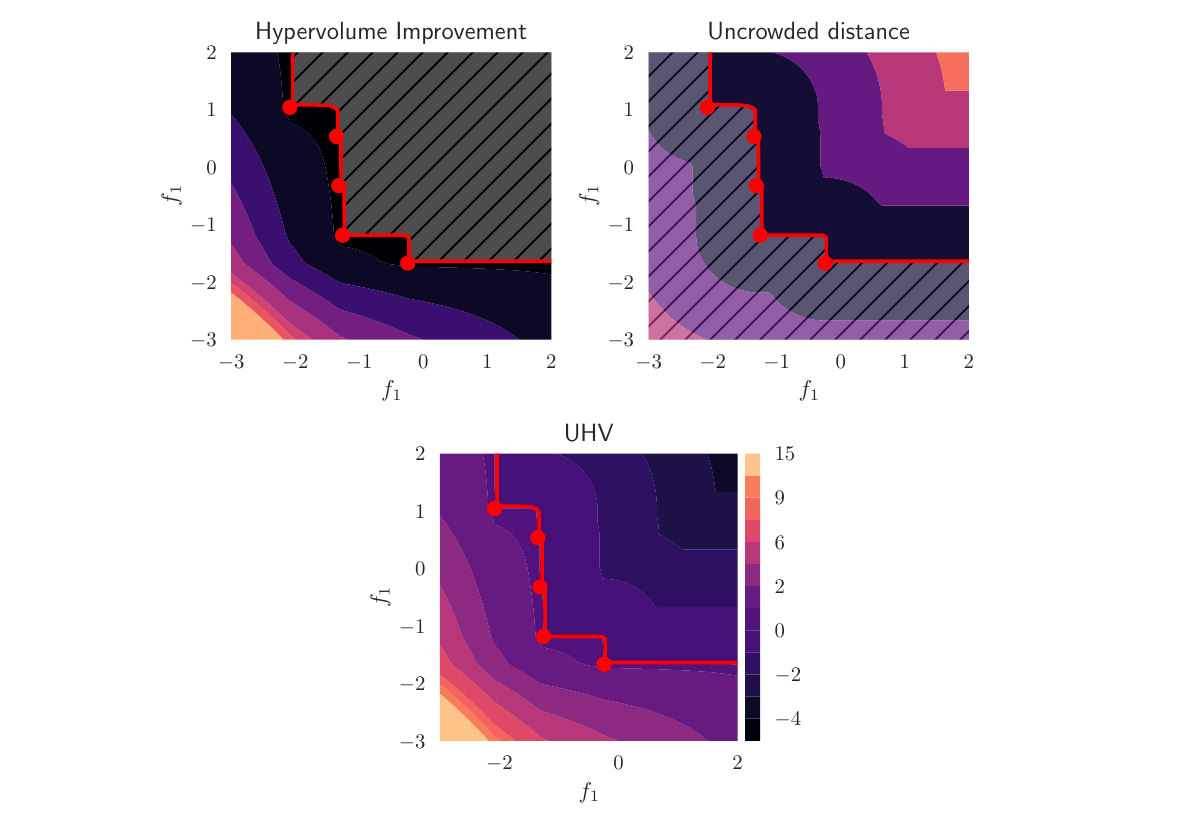}
    \caption{Illustration of the Uncrowded distance. The hatched regions correspond to the regions where the evaluated point does not contribute to the total loss.}
    \label{fig:HV_uncrowded_UHV}
\end{figure}

In order to nudge points toward the Pareto front, a modification of the
hypervolume, named the Uncrowded hypervolume (UHV) has been introduced in
\cite{toure_uncrowded_2019} in the context of genetic algorithms, and
further studied in gradient based optimization in
\cite{maree_uncrowded_2020,deist_multiobjective_2023}. We will first
consider the deterministic setting, where $\gb(\cdot)$ can be thought of
$\fb(\cdot, u)$ for $u\in \Uspace$. This modification takes the form of a
penalization as defined \cref{eq:def_uhv}, which is non-zero only for the points
which are dominated, and can be interpreted as a distance to the interpolated
empirical Pareto
front: 
\begin{align}
    \UHV(\Xoptim) &= \HV(\gb(\Xoptim)) - \lambda \sum_{x \in \Xoptim} d(\gb(x), \text{non-dom}(\gb(\Xoptim))) \label{eq:def_uhv}\\
    &= \HV(\gb(\Xoptim)) - \lambda \sum_{x \in \Xoptim} \inf_{p\in \mathrm{epf}(\gb(\Xoptim))} \|\gb(x)-p\|^2\, ,
\end{align}
where $\mathrm{epf}$ stands for (interpolated) empirical Pareto front and $\lambda > 0$ controls the impact of the penalization. It follows that if
all points in $\gb(\Xoptim)$ are non dominated, $\inf_{p\in
 \mathrm{epf}(\gb(\Xoptim))}\|\gb(x)-p\|^2 =0$. This is illustrated
\cref{fig:HV_uncrowded_UHV}, where the interpolated empirical Pareto front is
represented as the red broken line.

The extension to the stochastic case is direct, as $\UHV(\mathfrak{X}, U)$ is
now a random variable, and we can rewrite \cref{eq:bayes_expected_hv} using the
UHV. The optimization problem becomes
\begin{align}
         \min_{\substack{\mathfrak{X} \subset \Xspace\\|\mathfrak{X}| \leq n}}\mathcal{L}(\mathfrak{X}) \label{eq:stoch_uhv}\quad \text{with} \quad \mathcal{L}(\mathfrak{X}) = \Ex_U\left[\UHV(\mathfrak{X}, U)\right] \,.
\end{align}
\subsection{Gradient-based optimization of the mean UHV}
The stochastic optimization problem defined \cref{eq:stoch_uhv} can be solved
using now widespread stochastic optimization algorithm such as SGD, or Adam
\cite{wang_hypervolume_2017,ha_hybridizing_2022}, provided we can get an
unbiased estimator of the gradient of the loss defined~\cref{eq:stoch_uhv}.
In practice, this can be done using automatic differentiation.

The estimator of the gradient can be obtained in different ways, depending on the
application, and the nature of the uncertain space.

\paragraph{Deterministic optimization using Common Random Number}
We first consider the setting where we have a finite set of samples of the
uncertain variable $u_{\mathrm{crn}} = \{u^{(1)},\dots,u^{(N)}\}$.
\begin{align}
     &\min_{\substack{\mathfrak{X} \subset \Xspace\\|\mathfrak{X}| \leq n}}\mathcal{L}_{\mathrm{aao}}(\mathfrak{X}) \quad \text{with} \quad \mathcal{L}_{\mathrm{aao}}(\mathfrak{X}) = \frac{1}{N} \sum_{i=1}^N \UHV(\mathfrak{X}, u^{(i)})\,.
\end{align}
\paragraph{Stochastic Optimization}
The minibatches can be constructed by sampling a fixed number of uncertain variables:
\begin{equation}
    B = \{u^{(1)},\dots,u^{(|B|)}\}\text{ where }u^{(i)} \sim U\text{ iid}\,.
\end{equation}
In this case, the loss $\mathcal{L}_{\text{mb}}$ associated with the minibatch $B$ is defined
as
\begin{align}
     \min_{\substack{\mathfrak{X} \subset \Xspace\\|\mathfrak{X}| \leq n}}\mathcal{L}_{\mathrm{mb}}(\mathfrak{X}, B)\quad \text{with} \quad\mathcal{L}_{\mathrm{mb}}(\mathfrak{X}, B) = \frac{1}{|B|} \sum_{i\in B} \UHV(\mathfrak{X}, u^{(i)})\,.
\end{align}
\subsection{Illustration on BraninCurrin problem}

\begin{figure}[!ht]
    \centering
    \includegraphics[width=0.5\linewidth]{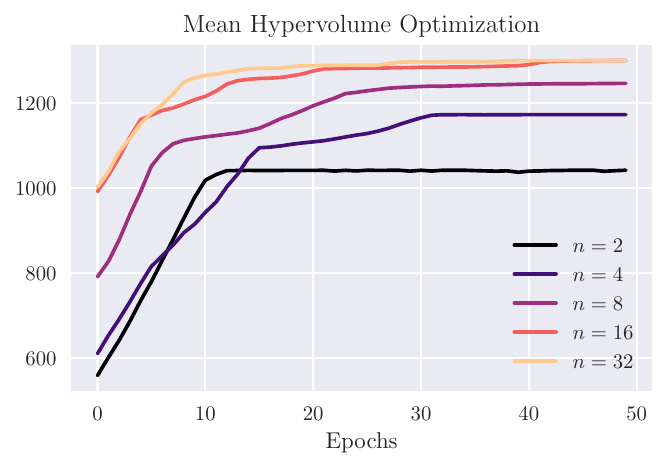}
    \caption{Estimation of the expected hypervolume depending on the number of points to consider in the optimization}
    \label{fig:inequality_gap_truth}
\end{figure}

We will illustrate some of the aforementioned properties on the BraninCurrin
function, as defined \cref{sec:branin_currin}. One straightforward observation
is that when the cardinality of the solution set $n$ increases, the maximal
value reached gets closer to the upper bound \cref{eq:hv_ub}, as illustrated
\cref{fig:inequality_gap_truth}.
When looking at the number of epochs needed to reach stable
results, we can see that for relatively large $n$, the optimization procedure
has less of an influence. This can be explained by the initial candidate, which is
sampled using Sobol' QMC method. This allows for good space fillings properties,
thus the set of points contains already a good coverage of the union of all
conditional Pareto sets.
To balance good coverage properties, interpretability,
and usability, we thus suggest using a moderate number of points $n$.

\begin{figure}[!ht]
    \centering
    \includegraphics[width=0.65\textwidth]{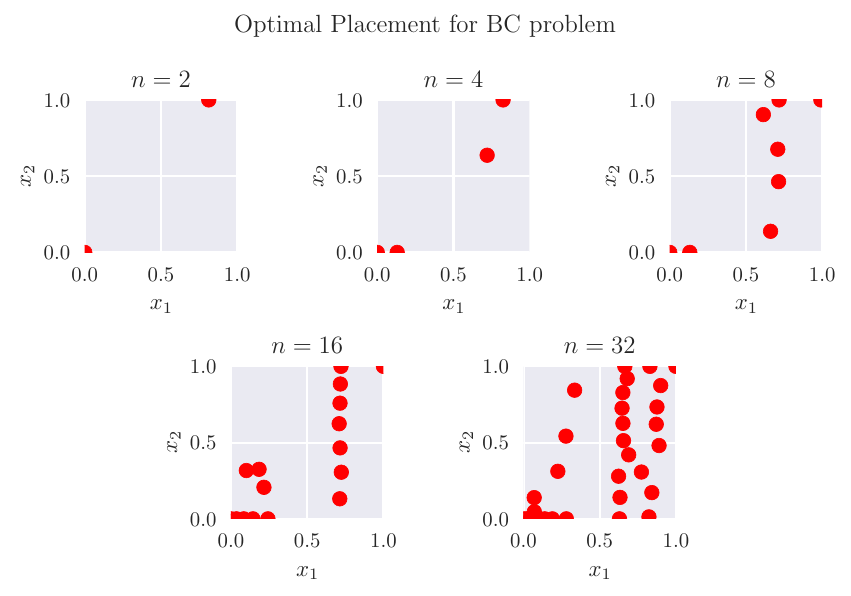}
    \caption{Optimizers of the UHV optimized using stochastic gradient for the BC problem}
    \label{fig:optimal_bc}
\end{figure}

\begin{figure}[b]
\centering
\begin{subfigure}{\textwidth}
    \centering
    \includegraphics[width=0.85\textwidth]{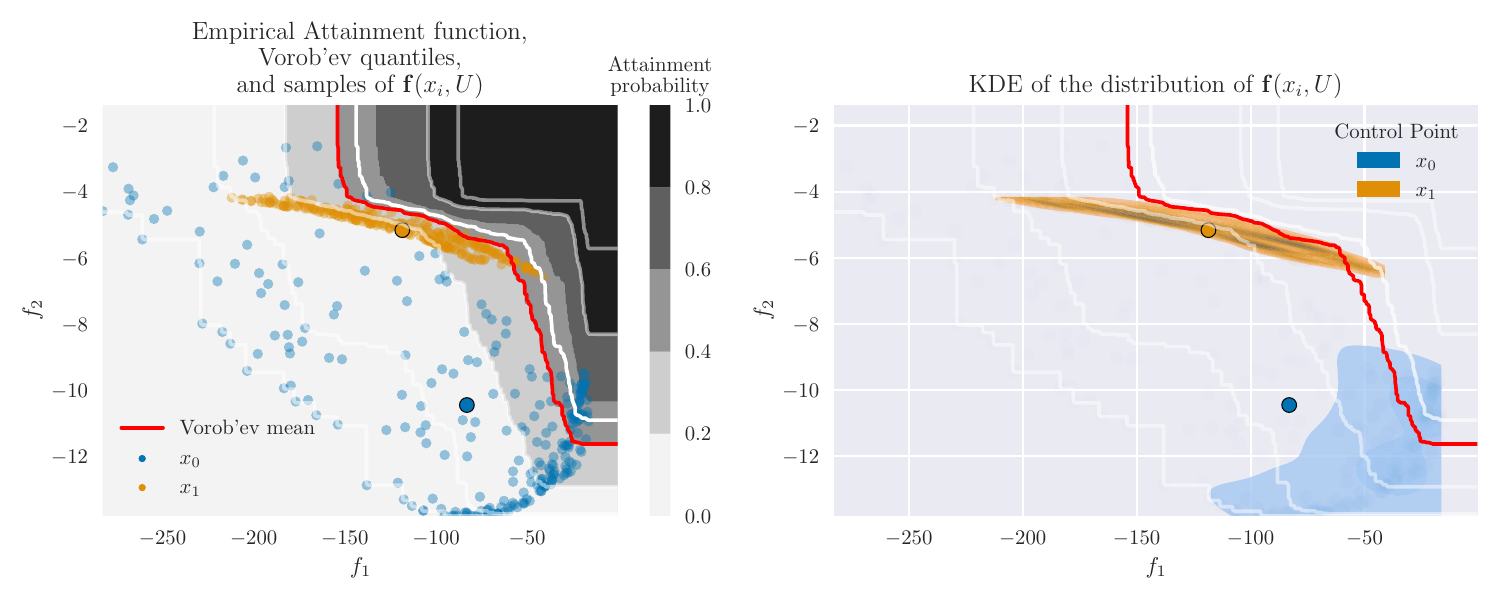}
    \caption{Distribution of $\fb(x_i, U)$ for $n=2$ considered in the optimization}
    \label{fig:kde_2}
\end{subfigure}
\begin{subfigure}{\textwidth}
    \centering
    \includegraphics[width=0.85\textwidth]{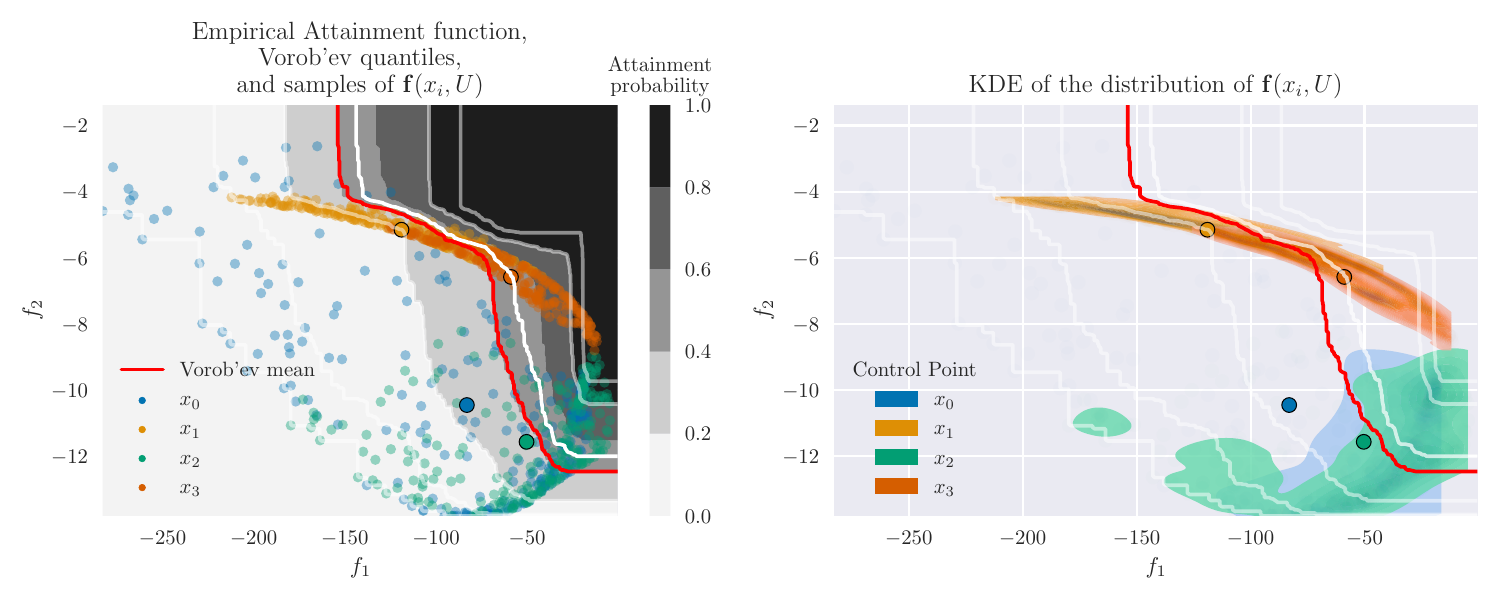}
    \caption{Distribution of $\fb(x_i, U)$ for $n=4$ considered in the optimization}
    \label{fig:kde_4}
\end{subfigure}
\begin{subfigure}{\textwidth}
    \centering
    \includegraphics[width=0.85\textwidth]{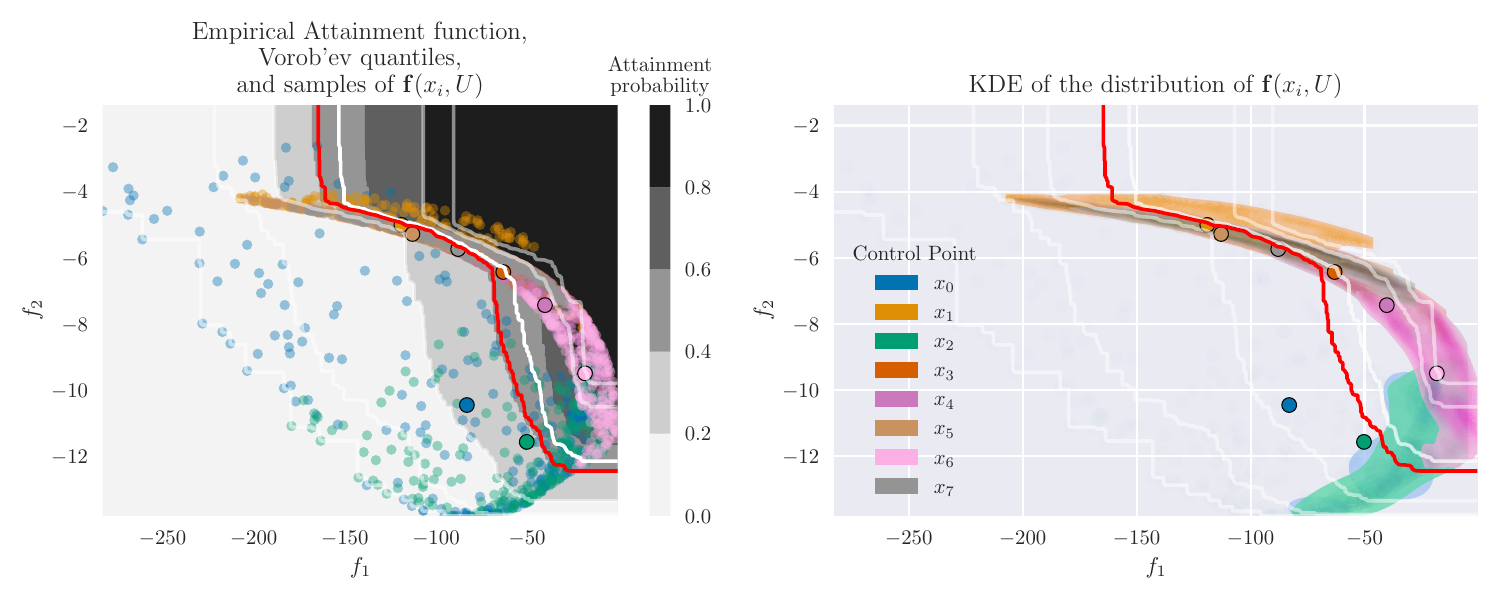}
    \caption{Distribution of $\fb(x_i, U)$ for $n=8$ considered in the optimization}
    \label{fig:kde_8}
\end{subfigure}%
\caption{Distribution of the solutions of the optimization problem for different $n$ on the BraninCurrin problem. The red line represents the Vorob'ev mean}
\label{fig:kde_bc}
\end{figure}

\Cref{fig:optimal_bc} shows the optimal values for the BraninCurrin problem,
optimized using the stochastic UHV. One can notice that the maximizers for $n'$
are included as maximizers for $n>n'$. Finally, \cref{fig:kde_bc} shows the
distribution of the $\fb(x_i, U)$ for $\{x_i\}_{1\leq i \leq n}$. One can see
that for the point $x_1$ on \cref{fig:kde_2,fig:kde_4}, its mean is
not completely representative of the distribution of $\fb(x_1, U)$, as it is
influenced by outliers.

\clearpage
\section{Gaussian Process Regression and Bayesian Optimization}
\label{sec:gp_bo}
In many practical cases, the gradient of $\fb$ with respect to the control
variable is not available directly, or requires a significant cost overhead.
This renders the gradient based optimization introduced above impractical to
implement. In order to reduce the overall computational cost, we developed a
method based on Gaussian Process and Active learning. The use of a
differentiable surrogate allows us to optimize the UHV in a time effective way,
while the Active Learning procedure improves the surrogate model
in a goal-oriented way.

\subsection{Gaussian Process regression in the joint space}
The objective function $\fb$ is modelled using a GP on the joint space
$\Xspace\times \Uspace$ by first specifying a GP prior, written $\Fb$, of
zero-mean and a covariance kernel, such as Matérn 5/2 (assuming twice
differentiability of $\fb$). We refer the reader to \cite{garnett_bayesian_2023}
for more details.

This GP prior is then conditioned using an initial design of experiment which
consists in a set of $N_0$ input/output pairs
$\mathcal{D}_0 = \left\{((x_1,u_1), \fb(x_1,u_1)),\dots,((x_{N_0},u_{N_0}),
\fb(x_{N_0},u_{N_0}))\right\}$.
This design is usually chosen as a space-filling design such as a Latin
Hypersquare. Conditioning $\Fb$ on this design of experiments gives
\begin{equation}
    \Fb \mid \mathcal{D}_0\sim \mathrm{GP}(\mb_\Fb, \mathbf{k}_{\Fb})\,,
\end{equation}
where $\mb_\Fb: \Xspace \times\Uspace \rightarrow \mathbb{R}^d$ is the kriging
mean, and $\mathbf{k}_\Fb: (\Xspace \times \Uspace) \times (\Xspace \times
\Uspace) \rightarrow \mathbb{R}^{d \times d}$ is the covariance function after
the conditioning. Based on this modelling of the unknown function, we can use
directly the kriging mean $\mb_\Fb$ as a proxy of the unknown function
$\fb$.

One significant
advantage of such a method is that the regularity of the surrogate is controlled
by the kernel $\mathbf{k}_\Fb$, meaning that the kriging mean is usually chosen
to be differentiable, and obtaining the gradient with respect to its input is
straightforward, either analytically, or using autodifferentiation.

\subsection{Active Learning}
\label{ssec:active_learning}
Based on the surrogate defined above, we can define an active learning
procedure. Starting from a training dataset (e.g. the design of experiment
$\mathcal{D}_0$) and a surrogate model constructed on it, the surrogate is used
to choose the next point that will be added to the design of experiment. More
specifically, in Bayesian Optimization, we use the properties of Gaussian
processes in order to select the next point to add. This point can be chosen in
a goal oriented way, for optimization as in the seminal work of
\cite{jones_efficient_1998} for instance. This is detailed in
\cref{alg:bayesopt}. In multiobjective settings, similar acquisition functions have
been developed, either as an extension of single objective optimization by means
of scalarization \cite{knowles_parego_2006}, but also directly using the notion
and properties of Pareto front
\cite{emmerich_single_2006,picheny_multiobjective_2013,zuluaga_active_2013,rojas-gonzalez_survey_2020},

\begin{algorithm}[ht]
\caption{Pseudocode of the typical Bayesian Optimization Loop}
\label{alg:bayesopt}
\begin{algorithmic}
\Procedure{BayesianOptimization}{$\mathcal{D}_0,\,\Fb,\,\text{budget}$} \State
\textbf{Input:} initial dataset
$\mathcal{D}_0=\{((x_i,u_i),\fb(x_i,u_i))\}_{i=1}^{N_0}$, GP prior $\Fb$, budget
\State $\mathcal{D}\gets\mathcal{D}_0$
\State $k\gets 0$ \While{budget not spent}
\State Condition $\Fb$ on $\mathcal{D}$ \Comment{Get the posterior GP
$\Fb\mid\mathcal{D}$}
\State Define acquisition using $\Fb \mid\mathcal{D}$
\State Use acquisition to get
$\displaystyle(x_{k+1},u_{k+1})$
\State $y_{k+1}\gets \fb(x_{k+1},u_{k+1})$ \Comment{Evaluate the
new point}
\State $\mathcal{D}\gets\mathcal{D}\cup\{((x_{k+1},u_{k+1}),y_{k+1})\}$ \Comment{Update
the DoE}
\State $k\gets k+1$ \EndWhile
\State \Return $\mathcal{D}, \Fb\mid
\mathcal{D}$
\EndProcedure
\end{algorithmic}
\end{algorithm}

\subsection{Lookahead hypervolume Improvement}
\label{sec:lookahead}
As we are looking for the set of $n$ points which maximizes the expected volume
of the region dominated, this objective can be rewritten using the GP prediction
as a plug-in estimation of the unknown function $\fb$:
\begin{equation}
    \max_{\substack{\Xoptim \subset \Xspace \\ |\Xoptim| \leq n}} \Ex_U\left[\HV\left(\mb_{\Fb}(\Xoptim, U)\right)\right]\,. \label{eq:max_e_hv_m}
\end{equation}

Writing the SUR criterion \cite{bect_supermartingale_2019} associated with
this optimization problem yields the following criterion in the joint space:
\begin{align}
    \alpha_{\text{SUR}}(x, u) = \Ex_{\Fb(x,u)}&\left[\max_{\substack{\Xoptim \subset \Xspace \\ |\Xoptim| \leq n}} \Ex_U\left[\HV\left(\mb_{\Fb \mid \mathcal{D}_{(x,u)}}(\Xoptim, U)\right)\right]\right] \label{eq:def_alpha_sur} - \max_{\substack{\Xoptim \subset \Xspace \\ |\Xoptim| \leq n}} \Ex_U\left[\HV\left(\mb_{\Fb}(\Xoptim, U)\right)\right] 
\end{align}
where  $\mathcal{D}_{(x,u)} = \mathcal{D}\cup\left\{((x, u), \Fb(x, u))\right\}$
is the design of experiments had we observed $\Fb(x, u)$ at $(x,u)$. This
criterion defined \cref{eq:def_alpha_sur} corresponds to the expected gain in
the maximal expected hypervolume, when adding the point $(x,u)$ to the design of
experiment.

However, as shown previously, solving the inner maximization problem of
\cref{eq:max_e_hv_m} is possible, but completely impractical in the context of a
SUR method, especially since the criterion $\alpha_{\text{SUR}}$ is supposed to
be optimized afterward. We propose instead a more tractable alternative, and to
consider \cref{eq:hv_ub} applied to $\mb_\Fb$
\begin{equation}
 \max_{\substack{\Xoptim \subset \Xspace \\ |\Xoptim| \leq n}}\Ex_U\left[\HV\left(\mb_{\Fb}(\Xoptim, U)\right)\right] \leq  \Ex_U\left[\HV\left(\mb_{\Fb}(\Xspace, U)\right)\right]
\end{equation}
In order to compute this quantity, we can then use a large number of candidate
points. The main difference being that instead of optimizing repeatedly the
expected hypervolume, one focuses on the upper-bound, which can be estimated using $X_{\text{est}}$, a
finite subset of $\Xspace$ with $|X_{\text{est}}|=N\gg n$

A more tractable acquisition function can then be defined using this upper bound:
\begin{equation}
    \tilde{\alpha}(x, u) = \Ex_{\Fb(x,u)}\left[\Ex_U\left[\HV\left(\mb_{\Fb \mid \mathcal{D}_{(x,u)}}(X_{\text{est}}, U)\right)\right]\right]\label{eq:def_alpha_tilde} - \Ex_U\left[\HV\left(\mb_{\Fb}(X_{\text{est}}, U)\right)\right]
\end{equation}
which represents the expected improvement of the upper bound once $(x,u)$ has
been added to the design. . Since computing analytically $\tilde{\alpha}$ is
untractable, let alone its gradient, we rely on Monte-Carlo methods in order to
approximate the expected values, by sampling 
\begin{itemize}
\item "fantasized" points, which correspond to samples of $\Fb(x, u)$, used to
approximate the outer expectation
\item a finite set $\Uspace_{\text{MC}}$ for the expectation with respect to $U$.
\end{itemize}

\subsection{Stochastic EHVI}
We also propose to consider a simple stochastic version of the EHVI. This
criterion is widely used in Bayesian Multiobjective Optimization, where it
represents the increase in the hypervolume of the currently non-dominated
points.
To adapt it to the stochastic case, we propose at iteration $k$ to sample first
an uncertain variable $u_k \sim U$, then to compute the EHVI conditioned on
$u_k$, that is $x \mapsto \mathrm{EHVI}(x, u_k)$ and to optimize it so that
$x_{k+1} \gets \argmax_{x}\mathrm{EHVI}(x, u_k)$.
\section{Numerical Results}
The optimization of the acquisition function can be a challenge in itself.
Indeed, they often rely on nested expectation, which have to be estimated with
sufficient precision. Some framework, such as Botorch
\cite{balandat_botorch_2020} which is based upon Torch, or Trieste
\cite{picheny_trieste_2023} based on Tensorflow aims at reducing the
computational overhead by using differentiable programming in order to compute
efficiently gradients.
\subsection{Experimental setting}
In the next sections, we compare numerically the different enrichment
strategies:
\begin{itemize}
    \item the Stochastic EHVI, labelled as $\mathrm{EHVI}$
    \item The lookahead hypervolume improvement as described
    \cref{sec:lookahead}, labelled $\mathrm{SUR}$
    \item A completely random strategy, labelled $\mathrm{rand}$, where the next
    point is chosen by sampling uniformly on the joint input space
    \item A version of the stochastic $\mathrm{EHVI}$, where instead of sampling
    the uncertainty, the value $\Ex[U]$ is chosen systematically, this leads to
    ignoring the uncertainty.
\end{itemize}
\subsection{4d problem}
For a first problem, we define a $\dim(\Xspace)=\dim(\Uspace)=2$ problem, based
on BraninCurrin problem (more details can be found in \cref{sec:branin_currin})
which leads to a GP surrogate on a 4-dimensional input space. As a rule of
thumb, we start by sampling $10$ times the input space dimension, so an initial
design of 40 points in $\Xspace\times\Uspace$, and add points until reaching 200
points.

\begin{figure}[!ht]
    \centering
    \includegraphics[scale=0.60]{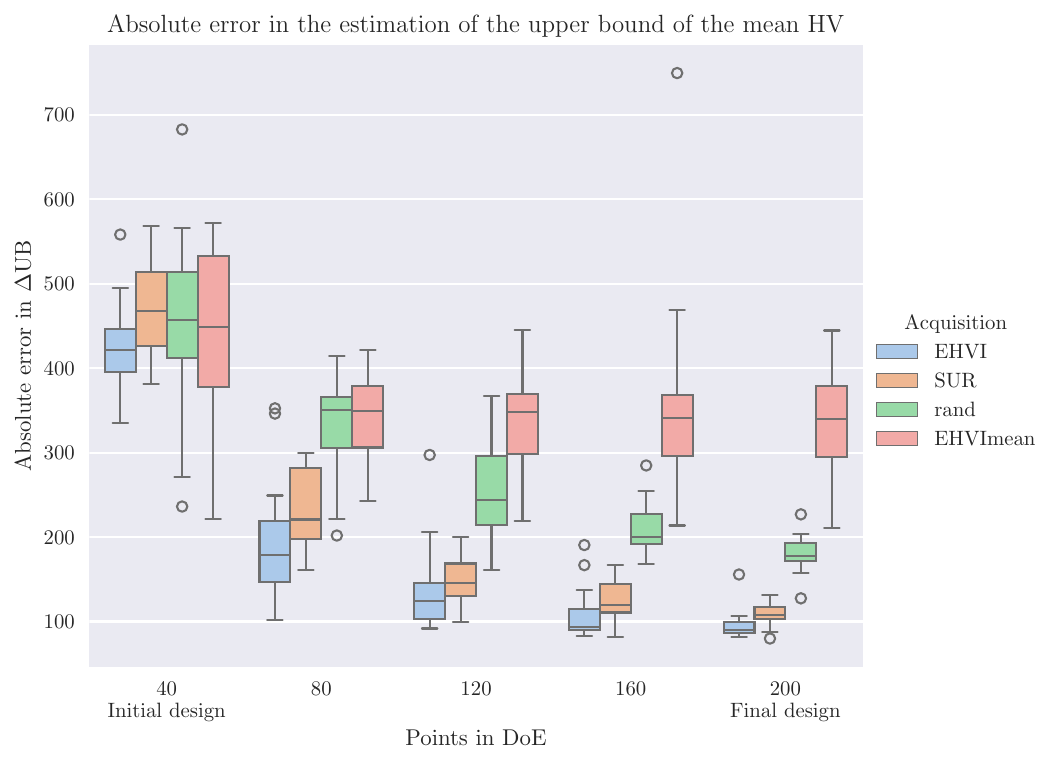}
    \caption{4DProblem: Error in the estimation of the upper bound, depending on the number of points added to the design of experiment}
    \label{fig:abs_error_4d}
\end{figure}

We will compare the value of the upper bound $\Ex_U[\HV(\fb(\Xspace, U))]$ with
the estimation using the GP prediction $\Ex_U[\HV(\mb_{\Fb}(\Xspace, U))]$. This
is illustrated \cref{fig:abs_error_4d}. We can see that the StochasticEHVI
method (labelled EHVI on the figure) and the SUR method outperform significantly
the random infill method, which consists in sampling independently points in the
joint space $\Xspace\times \Uspace$. It seems furthermore that the Stochastic
EHVI exhibits slightly better performances than SUR. This can maybe be
attributed to the successive approximation needed to compute $\tilde{\alpha}$
criterion. As expected, Since \text{EHVImean} completely ignores the uncertainty,
it does not lead to an exploration of the input space, and yields the worst results.

\begin{figure}[!ht]
    \centering
    \includegraphics[width=0.9\textwidth]{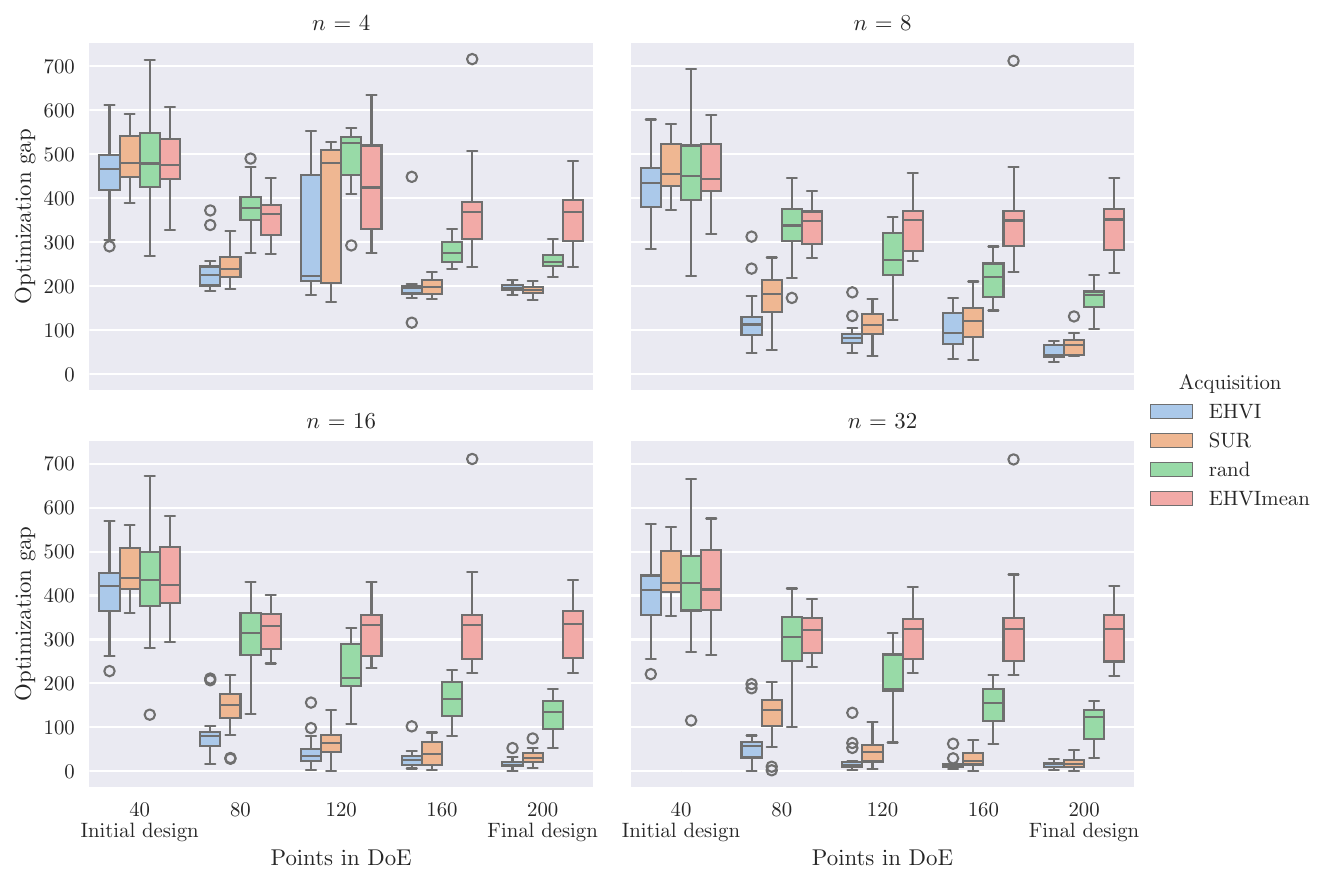}
    \caption{4DProblem: Optimization gap of the surrogate depending on the number of points considered for the optimization $n$, and the number of points in the design of experiments}
    \label{fig:optim_gap_4d}
\end{figure}

We now compare the value of the upper bound $\Ex_U[\HV(\fb(\Xspace, U))]$ with
the minimum obtained during the optimization procedure defined
\cref{eq:stoch_uhv}, where the UHV is computed with respect to the GP prediction
$\mb_{\Fb}$. This is illustrated \cref{fig:optim_gap_4d}. We can see that once
again, the StochasticEHVI and SUR methods outperform the random infill criterion. 

\clearpage
\subsection{10d problem}
We consider a problem where $\dim(\Xspace)=5=\dim(\Uspace)$, as in
\cite{trappler_multiobjective_2025}. Starting from an initial design of
experiments of 100 points, we added 400 points using the acquisition function
introduced \cref{eq:def_alpha_tilde}. As a baseline, we compare with a naive
strategy, where we add points randomly to the design. For each acquisition
function, we performed 16 replications. For the Bayesian Optimization loop, at
each iteration, we sampled randomly $|X_{\text{est}}| = 800$ points and
$|\Uspace_{\mathrm{MC}}|=32$ for the estimation of the upper bound. Each method
has been replicated 16 times. In order to compare the performances, we sampled
$|X^{(\text{test})}_{\text{est}}| = 10000$  and
$|\Uspace^{(\text{test})}_{\text{CRN}}| = 1024$ for the different estimations.
 
\begin{figure}[!ht]
    \centering
    \includegraphics[width=0.6\textwidth]{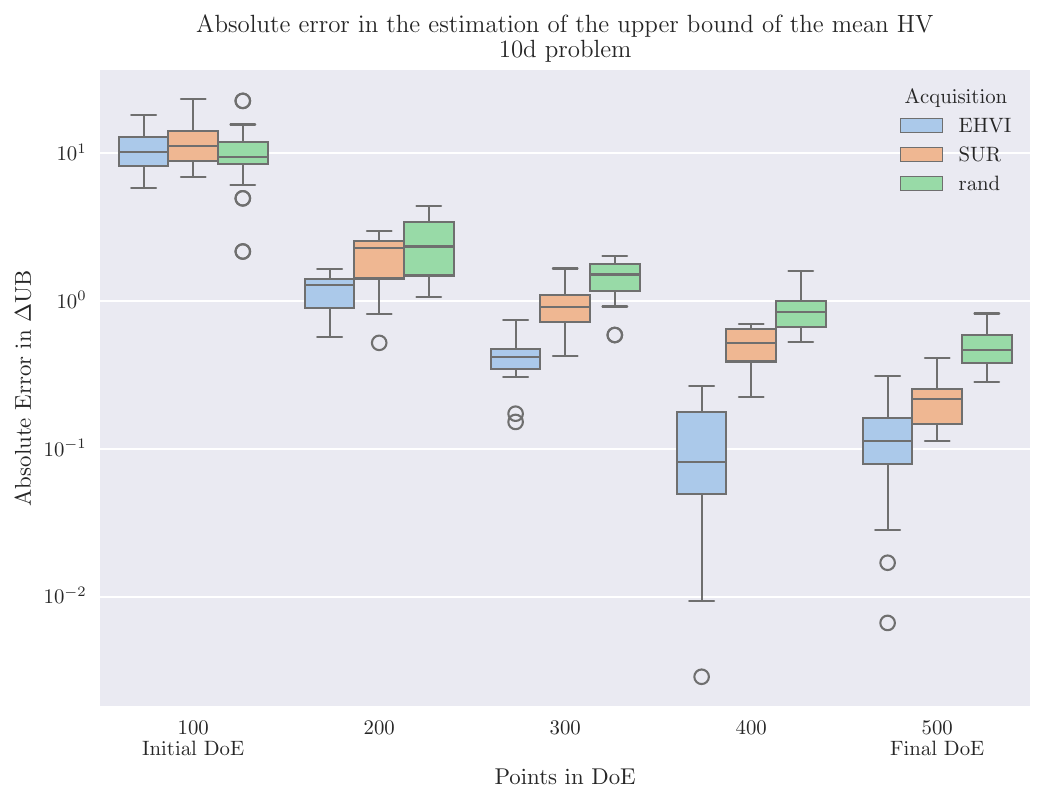}
    \caption{10d Problem: Error on the approximation of the upper bound}
    \label{fig:abs_error_10d}
\end{figure}

\begin{figure}[!ht]
    \centering
    \includegraphics[width=0.9\textwidth]{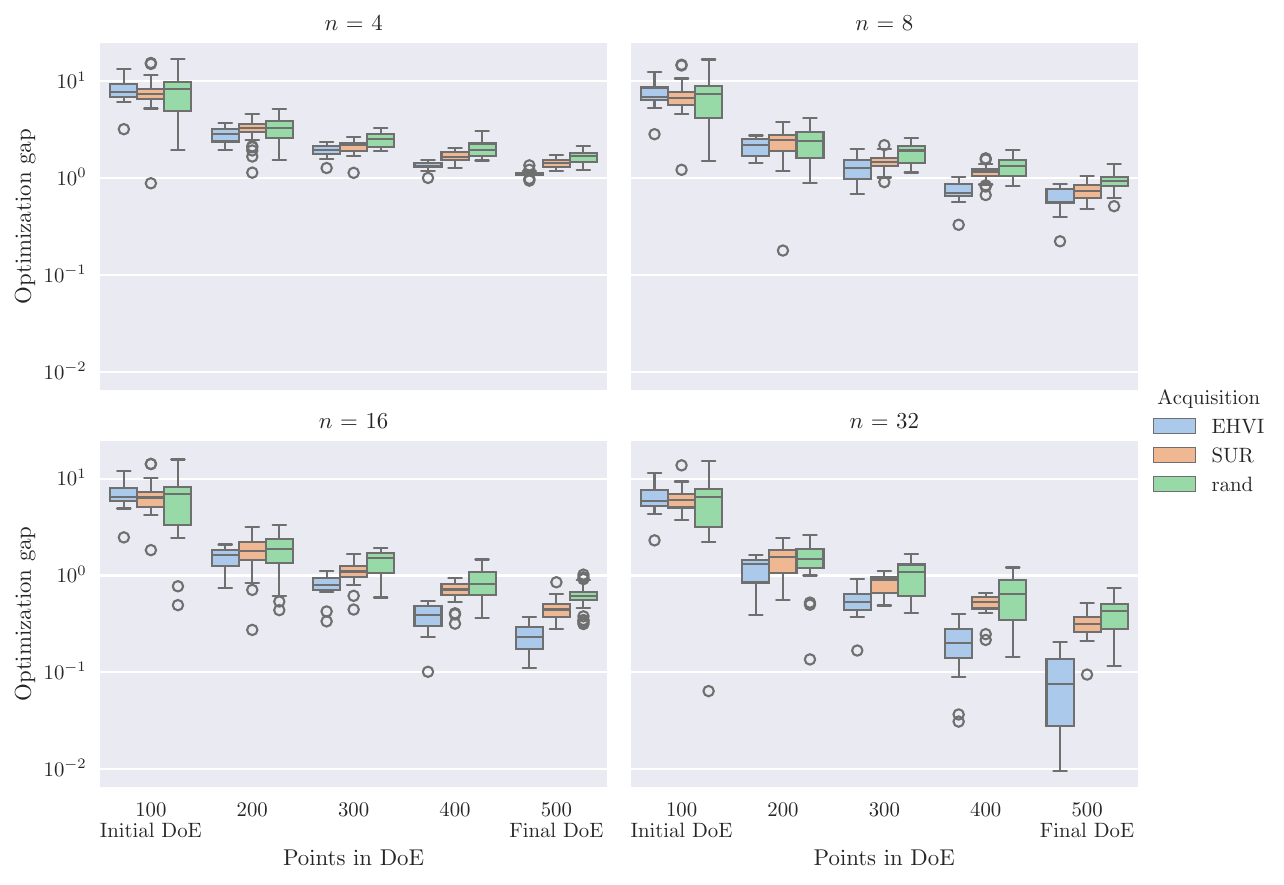}
    \caption{10d Problem: Optimization gap of the surrogate depending on the number of points considered for the optimization $n$, and the number of points in the design of experiments}
    \label{fig:optim_gap_10d}
\end{figure}

By design, the modified acquisition $\tilde{\alpha}$ aims at reducing the gap
between $\Ex_U\left[\max\HV(\mb_{\Fb})\right]$ and
$\Ex_U\left[\max\HV(\fb)\right]$. This is displayed \cref{fig:abs_error_10d} as
the distance between the quantities becomes smaller as the number of added
points increases. As in the previous lower dimensional example, StochasticEHVI
and SUR acquisitions perform better than random infill. The StochasticEHVI and the SUR methods also give significantly
better results in terms of optimization gap, as shown \cref{fig:optim_gap_10d}.

\clearpage

\section{Conclusion and discussion}
In this work, we present the Stochastic MOO problem as a stochastic optimization
problem with the use of the hypervolume, at the cost of increasing the dimension
of the optimization. In order to be able to solve efficiently this problem using
stochastic gradient-based methods, we used the UHV in order to nudge the
dominated points toward the Pareto front. As this optimization can be expensive
using directly the objective function, we propose to use a differentiable
surrogate model based on Gaussian Processes upon which we can perform the
procedure. Since the quality of the solution depends on the quality of the
approximation, we propose Active Learning methods based on the SUR framework,
and a stochastic version of the classical EHVI to add point sequentially to the
design of experiments in order to target specifically the regions of interest of
the original problem.

It also appears that optimization results for $n$ can be carried over for $n' >
n$. This could be further investigated in order to reduce the computational cost
of the optimization for large $n$, or give an alternative way to define an
acquisition function.

One shortcoming of the approach proposed in the SUR framework is the use of an
approximation of the upper bound for computational reasons, using $X_\text{est}$
in \cref{eq:def_alpha_tilde}. This approximation might explain the gap in
performance with the stochastic EHVI. Using a crude way to select a set of
points among $X_{\text{est}}$ which maximizes the original objective might
improve SUR method, or at least reduce the computational cost. Being able to
identify directly a subset of promising points, such as in
\cite{bringmann_twodimensional_2014}, or computing the expected hypervolume contribution
of every non-dominated point could be such ways to improve the acquisition function.
\section*{Acknowledgements}
This work was supported by the French National Research Agency (ANR) under the
France 2030 program, reference ANR-23-IACL-0006.

\printbibliography

\appendix

 \section{BraninCurrin}
 \label{sec:branin_currin}
\begin{figure}[!ht]
\centering
\begin{subfigure}{.5\textwidth}
    \centering
    \includegraphics[width=0.95\textwidth]{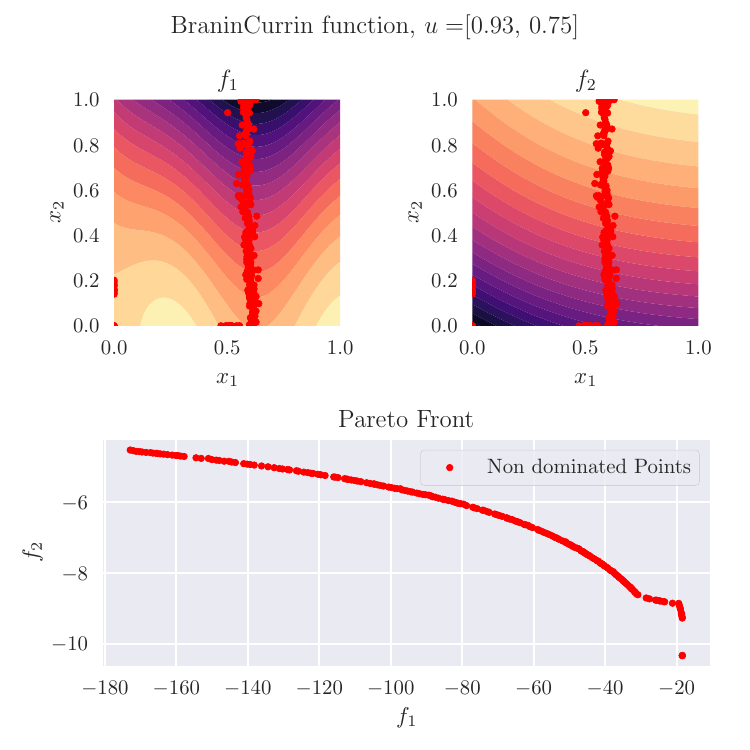}
\end{subfigure}%
\begin{subfigure}{.5\textwidth}
    \centering
    \includegraphics[width=0.95\textwidth]{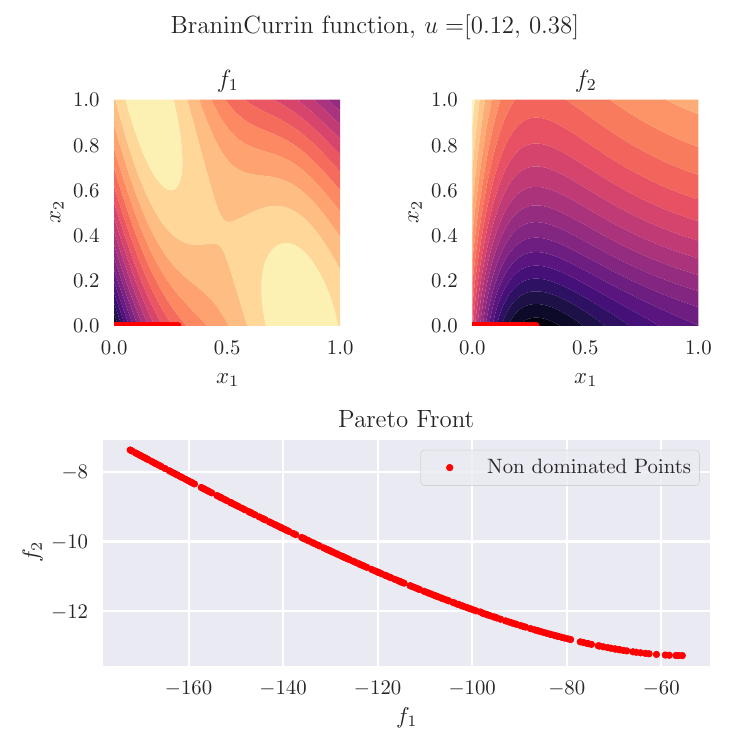}
\end{subfigure}
\begin{subfigure}{.5\textwidth}
    \centering
    \includegraphics[width=0.95\textwidth]{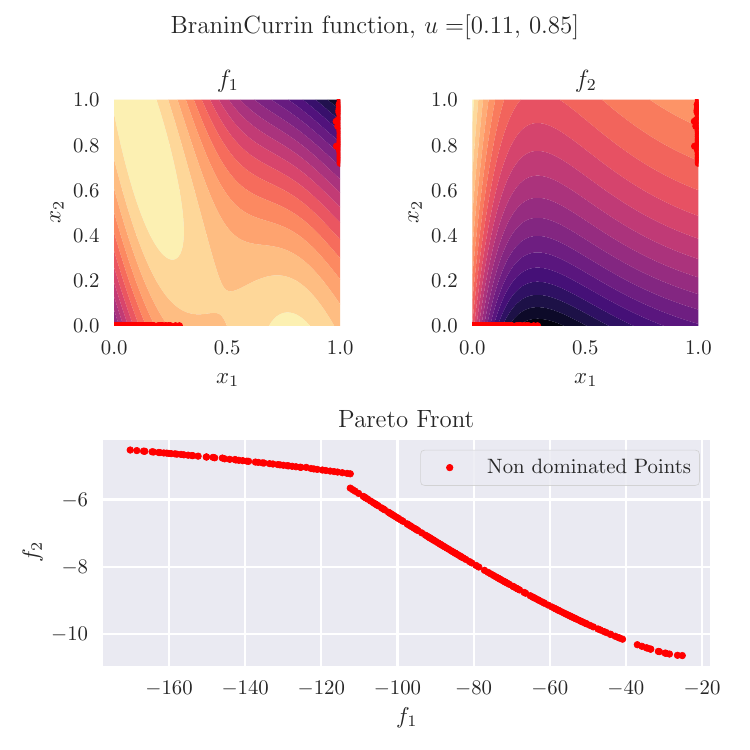}
\end{subfigure}%
\begin{subfigure}{.5\textwidth}
    \centering
    \includegraphics[width=0.95\textwidth]{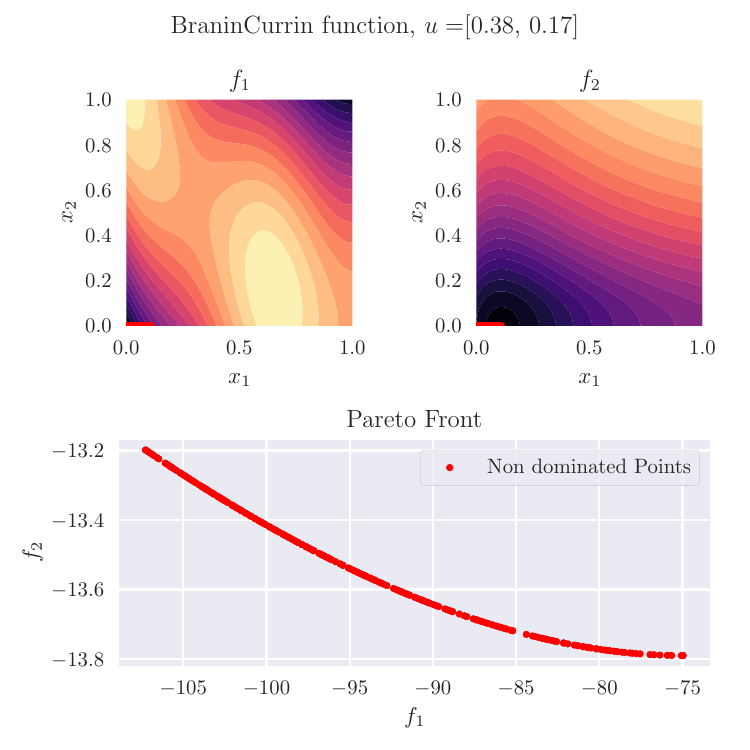}
\end{subfigure}
\caption{Solutions of BraninCurrin MO problem. Each subfigure corresponds to the problem conditioned on a different sample $u$. The red dots correspond to non-dominated points, which are different depending on $u$. }
\label{fig:illustration_bc}
\end{figure}

The BraninCurrin $\fb_{\mathrm{BC}}$ is composed of the two objectives:
\begin{align}
        f_1(x) &= \left(15x_2 - \frac{5.1}{4\pi^2} (15x_1 - 5)^2 + \frac{5}{\pi}(15x_1 - 5) - 5\right)^2 + \left(10 - \frac{10}{8 \pi}\right)\cos(15x_1 - 5)
  \\
 f_2(x) &= \left(1 - \exp\left(-\frac{1}{2x_2}\right)\right)\frac{
        2300 x_1^3 + 1900 x_1^2 + 2092 x_1 + 60}{100 x_1^3 + 500 x_1^2 + 4 x_1 + 20}
\end{align}
We introduce the uncertain variable as a perturbation of the input:
\begin{equation}
    \fb_{\mathrm{BC}}(x, u) = \left(-f_1(\mathrm{scale\_add}(x,u)), -f_2(\mathrm{scale\_add}(x,u))\right)
\end{equation}
where $\Xspace = [0, 1]^2$ and $\Uspace=[0,1]^2$, so that $U \sim
\mathrm{Unif}([0, 1])$
and
\begin{equation}
    \mathrm{scale\_add}(x,u) =x (1-\delta/2) + \delta + 2\delta(u-0.5)
\end{equation}
where the $\mathrm{scale\_add}$ function is introduced so that its output stays
in $[0, 1]^2$, and $\delta=0.2$.
\Cref{fig:illustration_bc} shows the problem for different sampled $u$ and the
non-dominated points found using NSGA-II, allowing us to illustrate the
influence of the uncertain variable on the Pareto Fronts and Pareto Sets.

\section{Interpretation as random sets}
Given the sets of points $\mathfrak{X}\in\Xspace^n$, we consider its hypervolume (with
respect to $\yb_{\text{ref}} = (r_1,\dots,r_d)$) as a random variable
$\HV(\fb(\mathfrak{X}, U))$. From another point of view, we can consider the
dominated region as a random closed set:
\begin{equation}
   D_\mathfrak{X}(U)= \bigcup_{x\in \mathfrak{X}} \prod_{i=1}^d [f_i(x, U), r_i]\,.
\end{equation}
The hypervolume is then the Lebesgue measure, noted $\mu$, of this random closed set. From the
theory of random sets, we can define the probability of coverage, or attainment
function \cite{dafonseca_attainmentfunction_2010} of $D_{\mathfrak{X}}$ as
\begin{equation}
    \begin{array}{rcl}
        p_{\mathfrak{X}}: \Xspace & \longrightarrow & [0, 1] \\
        x: & \longmapsto & \Prob_U[x \in D_{\mathfrak{X}}(U)] = \Prob_U\left[x\text{ is dominated by at least one point in } \fb(\mathfrak{X},U)\right] \,,
    \end{array}
\end{equation} and the Vorob'ev quantile \cite{molchanov_theory_2017} as $Q_{\alpha} = \{x \in \Xspace \mid p_\mathfrak{X}(x)\geq \alpha\}$.
The Vorob'ev mean of $D_{\mathfrak{X}}$ is a specific Vorob'ev quantile
$Q_{\alpha^*}$, which is defined as
\begin{equation}
    \mu(Q_{\alpha}) < \Ex_U\left[\mu(D_{\mathfrak{X}}(U))\right]\leq \mu(Q_{\alpha^*})\,,
\end{equation}
for any $\alpha > \alpha^*$. This can be interpreted as the Vorob'ev quantile
whose measure is the expected measure of the random closed set. 

Looking to optimize the mean hypervolume boils down to maximize the measure of
the Vorob'ev mean of the random dominated region. This approach using Vorob'ev
will be used as a visualization tools for the results, as shown
\cref{fig:kde_bc}.
\end{document}